\pdfoutput=1

\documentclass[11pt,a4paper]{article}

\usepackage{multirow}
\usepackage{subcaption}
\usepackage{amsmath}
\usepackage{subcaption}
\usepackage{graphicx,caption}
\usepackage{lscape}
\usepackage{linguex}
\usepackage{array}
\usepackage{booktabs}

\pdfoutput=1

\usepackage[]{EACL2023}

\usepackage[T1]{fontenc}

\usepackage{inconsolata}

\usepackage{tikz}
\usepackage{tikz-qtree}

\usepackage{cleveref}

\usepackage{times}
\usepackage{latexsym}
\usepackage[T1]{fontenc}

\usepackage[utf8]{inputenc}

\usepackage{microtype}
\title{Prosodic features improve sentence segmentation and parsing}

\author{
    {Elizabeth Nielsen~~~ Sharon Goldwater~~~ Mark Steedman}\\
    School of Informatics\\
    University of Edinburgh, UK \\
    {\tt\{e.nielsen, sgwater, steedman\}@ed.ac.uk} 
}  
  
\begin{document}
\maketitle
\begin{abstract}
Parsing spoken dialogue presents challenges that parsing text does not, including a lack of clear sentence boundaries.
We know from previous work that prosody helps in parsing single sentences \cite{tran2018}, but we want to show the effect of prosody on parsing speech that isn't segmented into sentences. 
In experiments on the English Switchboard corpus, we find prosody helps our model both with parsing and with accurately identifying sentence boundaries. 
However, we find that the best-performing parser is not necessarily the parser that produces the best sentence segmentation performance.
We suggest that the best parses instead come from modelling sentence boundaries jointly with other constituent boundaries.
\end{abstract}

\section{Introduction}

Parsing spoken dialogue poses unique difficulties, including speech disfluencies and a lack of defined sentence boundaries.
Because of these difficulties, current parsers struggle to accurately parse English speech transcripts, even when they handle other English text well.
Research has shown that prosody can improve parsing performance for speech that is  already divided into sentence-like units (SUs).\footnote{We follow \newcite{kahn2004} in using the term `sentence-like units' rather than `sentences' throughout, since conversational speech doesn't always consist of syntactically complete sentences.} \cite{tran2018,tran2019}.

In this work, we hypothesize that prosodic features from the speech signal will help with parsing speech that \textit{isn't} segmented into SUs, by improving the parser's ability to find SU boundaries.
We test this hypothesis by inputting entire dialog turns to a neural parser without SU boundaries.
These turns resemble the input a dialog agent would receive from a user.
We try two approaches: an end-to-end model that jointly segments and parses input, and a pipeline model that first segments and then parses the input.
To our knowledge, there hasn't been previous research on combining SU segmentation and parsing into a single task.
Following \newcite{tran2018,tran2019}, we consider two experimental conditions for each model: inputting text features only, and inputting both text and prosodic features extracted directly from the audio signal.
We also follow them in using the Switchboard corpus of English conversational dialogue.
Although overall parse scores are lower for parsers that don't have access to gold standard SU boundaries, our main hypothesis holds: that parsers using both text and prosodic features are more accurate than those using text alone. 
Unsurprisingly, the end-to-end model performs parsing better than the pipeline model because it doesn't suffer from error propagation.
We expected to find that gains in parsing quality would come primarily because models with access to prosody would perform SU segmentation better.
We do find that prosody helps all models improve their SU segmentation.
However, the pipeline model produces much better segmentation scores than the end-to-end model, and yet it still does worse at parsing.
In Section \ref{sec:results}, we discuss why segmentation and parsing quality do not always correlate in this task.
However, even though the best parses and segmentations are not always produced by the same model, all models perform better at both tasks with prosodic information.

Our primary contributions are:
\begin{itemize}
    \item We build an end-to-end model that jointly performs SU segmentation and parsing.
    \item We show that prosodic features are helpful for both SU segmentation and parsing, whether using an end-to-end or pipeline model.
    \item We show that an end-to-end model performs parsing better than a pipeline model, specifically because the end-to-end model is able to model SU boundaries jointly with other constituent boundaries.
\end{itemize}

\section{Background: prosody and syntax}

Prosodic signals divide speech into units \cite{pierrehumbert1980}.
The location and type of these prosodic units are determined by information structure \cite{steedman2000}, disfluencies \cite{shriberg2001}, and to some extent, syntax \cite{cutler1997}.
Some psycholinguistic research shows that in experimental conditions, speakers can use prosody to predict syntax (e.g., \newcite{kjelgaard1999}).
However, \newcite{cutler1997} argues that English speakers often ``fail to exploit'' this prosodic information even when it is present, so it isn't actually a signal for syntax in practice.
Many computational linguists have experimented with this possible link between syntax and prosody by incorporating prosody into syntactic parsers, which improves performance in some cases, but not all (e.g., \newcite{noeth2000,gregory2004,kahn2005,tran2018}).

Prosody's mixed record may be caused by the fact that prosodic units below the SU don't always coincide with traditional syntactic constituents \cite{selkirk1995,selkirk1984}. 
In fact, the only prosodic boundaries that consistently coincide with syntactic boundaries are the ends of SUs \cite{wagner2010}.
Prosodic boundaries at the end of SUs are more distinctive, with longer pauses and more distinctive pitch and intensity variations, making prosody a reliable signal for SU boundaries, even though it's less helpful for lower-level syntactic structure.

Some researchers have used prosody to help in SU boundary detection. 
Examples of SU segmentation models that benefit from prosody include \newcite{gotoh2000,kolar2006,kahn2004,kahn2012}, who all used traditional statistical models (e.g., HMMs, finite state machines, and decision trees), and \newcite{xu2014}, who used a neural model.

\section{Task and data}
\label{sec:task}

We use the American English corpus Switchboard NXT (henceforth SWBD-NXT) \cite{calhounetal2010} to allow us to compare performance with \newcite{tran2018} and \newcite{tran2019}.
SWBD-NXT comprises 642 telephone dialogues between strangers.\footnote{This corpus is relatively small compared to many speech datasets used today, but it is the largest speech corpus we know of with hand-annotated constituency parses.
While other corpora with hand-annotated dependency parses exist (e.g., UD corpora such as Wong et al. (2017) and Dobrovoljc and Nivre (2016)), these are all significantly smaller than Switchboard NXT.}
These dialogues are transcribed and hand-annotated with Penn Treebank-style constituency parses, and have no punctuation.
For acoustic features, we extract features for pitch, intensity, pause duration, and word duration from the audio signal, largely following the feature extraction procedure of \newcite{tran2018}, summarized in Appendix \ref{app:feats}.

The transcript divides the corpus into \textit{SUs} and \textit{turns}.
Since not all utterances are full sentences, we use the generic term `sentence-like unit' (SU).
A turn is a contiguous span of speech by a single speaker.
Not all turns in the SWBD-NXT contain multiple SUs: of a total 60.1k turns, 35.8k consist of a single SU.
The average number of SUs per turn is 1.82.

We follow the general approach of \newcite{tran2018}, but where they parse a single SU at a time, we parse a  whole turn.
We approach this task in two ways: an end-to-end model (SU-segmentation and parsing done jointly) and a pipeline model (SU-segmentation done before parsing).
Both models return constituency parses for each turn in the form of Penn Treebank (PTB)-style trees. 
In order to keep the output in the form of valid PTB trees for the end-to-end-model, we add a top-level constituent, labelled \textsc{turn}, to all turns, however many SUs they consist of. 
As we discuss in Section \ref{sec:conclusion}, this innovation allows the end-to-end model to treat SUs in the same way that it treats other syntactic units.


%
To avoid memory problems from too-long inputs, we filter out two problematically long turns from the training set (out of 49,294 turns).
We do not have to remove any turns from the development or test sets.
This leaves the maximum turn length at 270 tokens.
We also remove any turns for which some or all of the audio is missing.

\section{Model} \label{sec:model}

For both the end-to-end model and the pipeline parser, we use \newcite{tran2019}'s parser, extending the code base described in their paper.\footnote{Original: \href{https://github.com/trangham283/prosody_nlp}{https://github.com/trangham283/prosody\_nlp}; our extended code: \href{https://github.com/ekayen/prosody_nlp}{https://github.com/ekayen/prosody\_nlp}.}
The model is a neural constituency parser based on \newcite{kitaev2018}'s text-only parser, with a transformer-based encoder and a chart-style decoder based on \newcite{stern2017} and \newcite{gaddy2018}. 
This encoder-decoder is augmented with a CNN on the input side that handles prosodic features \cite{tran2019}.
For further description of the model and hyperparameters, see Appendices \ref{app:model} and \ref{app:hparams}.
The text is encoded using 300-dimensional GloVe embeddings \cite{glove}. 

For the pipeline, we first segment into SUs, and then parse the resulting SUs.
For segmentation, we use a modified version of the parser: With the same encoder, we change the decoder to only do sequence labeling, marking tokens as either SU-internal or SU-final.
We then parse the predicted SUs.
Rather than using a parser that was trained on gold SUs, we train a parser on the SUs that the segmenter predicted on the train set. 
This allows the model to learn to produce the parses on imperfectly segmented SUs and leads to better parsing scores.

We report two metrics for both the pipeline and end-to-end models: parse and SU segmentation F1 scores.
Parse F1 is calculated on the whole turn using a Python implementation of EVALB.\footnote{\href{https://github.com/ekayen/PYEVALB}{https://github.com/ekayen/PYEVALB}}
We don't count \textsc{turn} constituents, so that turn-based and SU-based parse scores are comparable.
The SU segmentation F1 score is calculated on all turn-medial SU boundaries; turn-final SU boundaries are not counted.
In order to calculate the SU segmentation F1 score for the end-to-end model, we consider every node that is a direct child of the tree's top \textsc{turn} node to be an SU.
That is, SUs are just one kind of syntactic constituent, differentiated only by their location in the tree.

Unless stated otherwise, we train each model on five random seeds, and report the mean of each metric over all five seeds.
To determine the statistical significance of differences between model performance, we use bootstrap resampling \cite{efron1994}, resampling $10^{5}$ times.

\section{Results and discussion} \label{sec:results}

\begin{table}
     \begin{tabular}{lccc} 
     \hline   
     & \textbf{Gold SUs}& \textbf{E2E} & \textbf{Pipeline}
     \\ [0.5ex] 
    \hline
    \textbf{Dev.\ set:} && \\
    Text only  & $90.31$ & $85.70$ & $84.34$ \\ 
     Text+prosody & $90.90$ & $86.21$ & $85.28$  \\ 
    \textbf{Test set:} && \\
    Text only  & $90.29$ & $86.03$ & $84.68$ \\ 
     Text+prosody & $90.65$ & $86.55$ & $85.62$  \\ 
     \hline

    \end{tabular}      

    \caption{
    Development and test set parsing F1 score of the end-to-end and pipeline models (and for comparison, a model that receives gold standard SUs as input). Results averaged over 5 random seeds.} 
\label{tab:parseresults}
\centering
\end{table}

     %
%

\begin{table}
\centering
     \begin{tabular}{llcc } 
     \hline 
&&  \textbf{E2E} & \textbf{Pipeline}
     \\ [0.5ex] 
    \hline
    \textbf{Dev.\ set:} && & \\
    Text only  & F1 & $66.32$ & $63.74$ \\ 
     \multirow{3}{*}{Text+prosody $\left\lbrace\begin{array}{l} \\ \\ \end{array}\right.$}   &  F1  & 
                                                       
                                    \multirow{3}{*}{  $\begin{array}{l}
                                                       72.95  \\
                                                       69.46 \\
                                                       76.92 \end{array} $} &
                                    \multirow{3}{*}{ $\begin{array}{l}
                                                        77.38 \\
                                                       79.44 \\
                                                       75.69 \end{array}$}\\ 

& Prec  && \\
& Rec && \\
     
\textbf{Test set:} &&& \\
    Text only   & F1 & $71.01$ & $66.98$ \\ 
     Text+prosody  & F1 & $72.94$ &  $77.38$  \\ 
     \hline
    \end{tabular}      

    \caption{
    Segmentation F1 score of the turn-based models compared to the SU-based model, with precision and recall given for some cases, averaged over 5 random seeds.
    } 
\label{tab:segresults}
\centering
\end{table}

Experiments with the end-to-end model show that prosody shows a statistically significant effect on parsing performance, though this effect is small (see Table \ref{tab:parseresults}). 
In the pipeline model, the effect of prosody is larger: The difference in parse F1 score from adding prosody is 0.94 for the pipeline model, where it is only 0.52 for the end-to-end model. 
However, the pipeline model's parse F1 score is lower than the end-to-end model's.

It's not surprising that the end-to-end model does better at parsing, since we expect errors to propagate from the segmentation step to the parsing step in the pipeline model.
What is surprising is that the pipeline model has a much higher segmentation score than the end-to-end model, which complicates the error propagation account.

We can explain this discrepancy by the kinds of errors each model makes.
First, the end-to-end model tends not to predict top nodes for each gold SU, instead connecting lower nodes in the tree directly to the \textsc{turn} node, leading to oversegmentation, as shown in Figure \ref{fig:decorrel}\subref{fig:gold}.
In Appendix \ref{app:dummywt} we describe how we discouraged oversegmentation, but the end-to-end model still tends to oversegment severely.
The pipeline model tends instead to undersegment.
This tendency is reflected in the end-to-end and the pipeline models' similar segmentation recall, compared to the end-to-end model's very low precision, shown in Table \ref{tab:segresults}. 
The examples in Figure \ref{fig:decorrel} show the effect this has on segmentation: The end-to-end example shown here has a segmentation F1 score 57.1, where the pipeline has a score of 66.7.

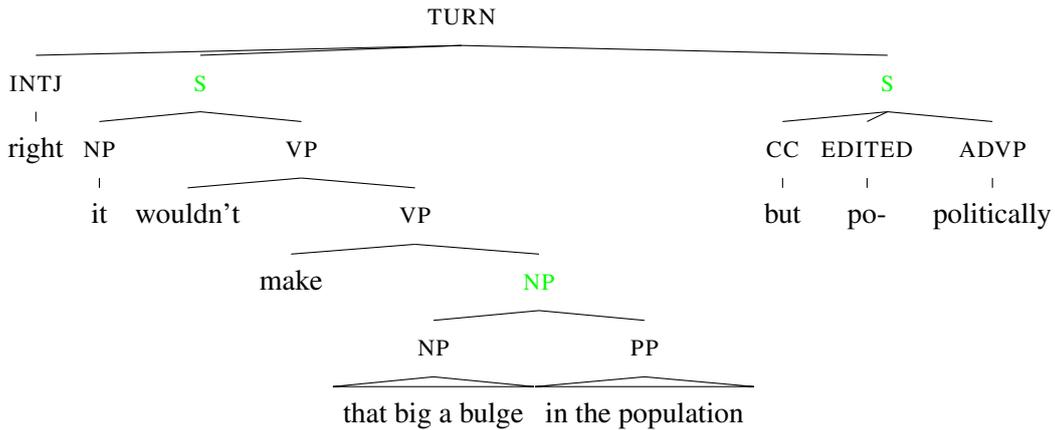
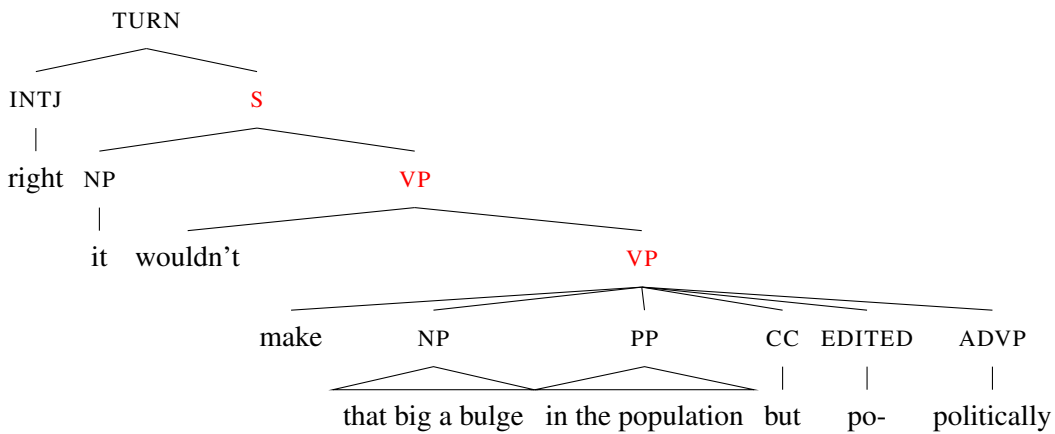
\begin{figure*}[ht]
    \centering
\begin{subfigure}{\textwidth}
 \centering 
    \begin{tikzpicture}
    \tikzset{level distance=25pt,sibling distance=0pt}
    \tikzset{execute at begin node=\strut}
    \Tree [.\textsc{turn} [.\textsc{intj} right ] [.\textsc{\textcolor{green}{s}}  [.\textsc{np} it ]  [.\textsc{vp} wouldn't  [.\textsc{vp} make  [.\textsc{\textcolor{green}{np}} [.\textsc{np} \edge[roof]; {that big a bulge} ] [.\textsc{pp} \edge[roof]; {in the population} ]  ] ] ] ]  [.\textsc{\textcolor{green}{s}} [.\textsc{cc} but ]  [.\textsc{edited}  po- ]  [.\textsc{advp}  politically  ] ] ] 
    \end{tikzpicture}
    \caption{Gold parse. The nodes shown in green are omitted by both the end-to-end and pipeline models.}       \label{fig:gold} 
\end{subfigure}
\begin{subfigure}{\textwidth}
 \centering
    \begin{tikzpicture}
    \tikzset{level distance=30pt,sibling distance=0pt}
    \tikzset{execute at begin node=\strut}
    \Tree [.\textsc{turn}  [.\textsc{intj} right ] [.\textsc{\textcolor{red}{s}} [.\textsc{np} it ] [.\textsc{\textcolor{red}{vp}} wouldn't  [.\textsc{\textcolor{red}{vp}}  make  [.\textsc{np} \edge[roof]; {that big a bulge} ] [.\textsc{pp} \edge[roof]; {in the population} ] [.\textsc{cc} but ] [.\textsc{edited} po- ] [.\textsc{advp} politically ] ] ] ] ] ] ] ] 
    \end{tikzpicture}
    \caption{Parse predicted by the pipeline model. The nodes shown in red are incorrect.} \label{fig:pipe}
\end{subfigure} 
      \caption{Comparison of gold parse to the pipeline's predicted tree. The end-to-end model's tree isn't shown because the only differences between it and the gold parse is the omission of the three nodes that are highlighted in green in the gold parse. This example has been edited for length and clarity, and part-of-speech tags have been omitted.} \label{fig:decorrel}
\end{figure*} 

However, when scoring the parses, the end-to-end model is penalized much less. 
Both the end-to-end and pipeline models omit three nodes from the tree in Figure \ref{fig:decorrel}.
However, the pipeline model also predicts several nodes high up in the tree that the end-to-end model does not.
This leads to a much lower parse F1 score for the pipeline model: 69.2, compared to the end-to-end model's 88.0.

In fact, the pipeline's better segmentation seems to actively \textit{worsen} its parse score. 
The pipeline's undersegmentation comes from predicting nodes that dominate entire predicted SUs\,---\,like the \textsc{s} node in Figure \ref{fig:decorrel}\subref{fig:pipe}.
Because the \textsc{s} node erroneously spans the entire turn, it is far more likely that its \textsc{vp} daughter will also span too many nodes, as it does in this example. 
In Appendix \ref{app:pipe}, we discuss measures we took to reduce this kind of error propagation, none of which eliminated this problem.


In order to demonstrate how the phenomena shown in this example affect the performance on the whole development set, we look at the overall interaction of parse precision and segmentation accuracy.
On examples where the SU boundaries are incorrectly predicted, as in the example in Figure \ref{fig:decorrel}, the pipeline model predicts many more incorrect nodes, and its parse precision declines by 5.22\% on the development set (from 85.72 for all parses to 80.50 for parses with incorrect SU boundaries). 
By comparison, the end-to-end model’s parse precision is less affected by segmentation quality – it only drops by 3.96 \% (from 86.20 to 82.24).

These results suggest that the end-to-end model's superior parsing ability comes from the fact that it is able to model all syntactic units, including SUs, in the same way.
The pipeline model has to treat SUs as being logically prior to and distinct from all sub-SU units, which leads to the error propagation described above.
By modeling SUs and other syntactic constituents similarly, the end-to-end model is able to propose the sub-SU nodes that lead to the best parses overall, without being bound to a certain SU segmentation.

\section{Conclusion} \label{sec:conclusion}

Previous work has shown that prosody improves parse quality.
In this work, we show that prosody improves parse and SU segmentation quality simultaneously. 
We show that parse and SU segmentation score are not necessarily correlated: A pipeline model does SU segmentation better, but an end-to-end model produces better parses.
We propose that this is because the end-to-end models SU boundaries the same way as from other syntactic boundaries.
By treating SUs as just another kind of syntactic unit, our model is able to take advantage of prosody to produce better parses overall.

\section{Limitations} \label{sec:limit}

A primary limitation of this work is the data used.
We chose to use Switchboard-NXT because it is one of a very few speech corpora with gold constituency parses available, and it is used in the work that is most directly comparable to ours.
Switchboard-NXT includes only North American English, and was specifically recorded in the early 1990s. 
We therefore don't know the extent to which our results hold for other varieties of English or for other languages.

Another limitation of this corpus is its size and quality.
Switchboard-NXT is relatively small (the section we used was approximately 50k turns), due to the difficulty of producing the thorough annotations this corpus includes.
The audio quality is also limited due to the age of the data: The original Switchboard recordings have a sample rate of just 8 kHz (compare to standard compact disk sample rates of 44.1kHz).
This reduces the resolution of acoustic features, and particularly affects high frequencies.
This could affect the quality of the pitch feature in particular. 
We would strongly prefer more and better quality data; unfortunately, collecting and annotating a comparable corpus would be prohibitively expensive.

Some of our conclusions about model behavior may be somewhat contingent on architecture.
In particular, the end-to-end model's tendency to oversegment might not be as severe in other parsers. 
We chose to use \citet{kitaev2018}'s parser in order to facilitate comparisons with the work of \citet{tran2019}, but this parser is unusual in that there is no direct modeling of the relationship between parent and child nodes, which is part of how what makes it an especially efficient parser. 
This means that the model doesn't directly penalize things like connecting leaf nodes to a top-level \textsc{turn} node, which is seen in many cases of oversegmentation by the end-to-end model.
If there were a mechanism for incorporating parent-child relationships between nodes into the loss, then it's quite possible that oversegmentation would be less pronounced in the end-to-end model.
Experimenting with different parser architectures with this goal is a possible direction for future work.

\section{Ethical statement}

In Sections \ref{sec:task} and \ref{sec:limit}, we describe the demographic composition of our dataset in as much detail as we have access to (noting that it contains North American English of the 1990s). 
It is likely that the system as we develop it here would not generalize consistently to speakers of other varieties of English.

We anticipate that a parser like ours, which handles data that doesn't have gold SU boundaries, would be most likely deployed in spoken dialog systems.
Our innovations are very unlikely to increase the harms that a spoken dialog system already poses (which include issues such as not being accessible to speakers of marginalized language varieties and potentially compromising user privacy).

\bibliography{acl2020,anthology}
\bibliographystyle{acl_natbib}
\appendix

\section{Appendix}

\subsection{Acoustic feature extraction} \label{app:feats}

Following \citealt{tran2018} and \citealt{tran2019}, we extract features for pauses between words, word duration, pitch, and intensity. 
We also experimented with adding features corresponding to voicing, but didn't see any benefit from this addition, and so we don't report these results here.

\textbf{Pause} features are extracted from the time-aligned transcript. 
Each word's pause feature corresponds to the pause follows it.
Each pause is categorized into one of six bins by length in seconds: $p > 1$, $0.2 < p \leq 1$, $0.05 < p \leq 0.2$, $0 < p \leq  0.05$, $p \leq 0$ (see below), and pauses where we are missing time-aligned data.
Following \newcite{tran2018}, the model learns 32-dimensional embeddings for each pause category.

Since we use turns instead of SUs, we have to determine how to handle pauses at the beginnings and endings of turns.
We decide to calculate pauses based on all words in the transcript, not just the words for a single speaker at a time. 
This means that at a turn boundary, we calculate the pause as the time between the end of one speaker's turn and the beginning of the other speaker's turn. 
If one speaker interrupts another, the pause duration has a negative value.
We place these negative-valued pauses in the same bin as pauses with length 0.

\textbf{Duration} features are also extracted from the time-aligned transcript. 
We are interested in the relative lengthening or shortening of word tokens, so we normalize the raw duration of each token. 
Following the code base for \newcite{tran2019}, we perform two different types of normalization.
In the first case, we normalize the token's raw duration by the mean duration of every instance of that word type. 
In the second, we normalize the token's raw duration by the maximum duration of any word in the input unit (SU or turn). 
These two normalization methods result in two duration features for each word token, which are concatenated and input to the model.

\textbf{Pitch} features (or more accurately, F0 features) are extracted from the speech signal using Kaldi \cite{povey2011}. These are extracted from 25ms frames every 10ms. Three pitch features are extracted: warped Normalized Cross Correlation Function (NCCF); log-pitch with mean subtraction over a 1.5-second window, weighted by Probability of Voicing (POV); and the estimated derivative of the raw log pitch. For further details on these features, see \newcite{Ghahremani2014}.

\textbf{Intensity} features are also extracted from the speech signal using the same software and frame size as we use for pitch features. Starting with 40-dimensional mel-frequency filterbank features, we calculate three features: (1) the log of the total energy, normalized by the maximum total energy for the speaker over the course of the dialog; (2) the log of the total energy in the lower half of the 40 mel-frequency bands, normalized by the total energy; and (3) the log of the total energy in the upper half of the 40 mel-frequency bands, normalized by the total energy.

For training, development, and testing, we use the split described in \newcite{charniak2001}, which is a standard split for experiments on SWBD-NXT (e.g., \newcite{kahn2005,tran2018}). 
The training set makes up 90 percent of the data, and the development and testing sets make up 5 percent each.

\subsection{Model description} \label{app:model}

The parser is an encoder-decoder model that takes both speech and text inputs. In this appendix, we describe the three main model components: the CNN that processes the continuous speech inputs before they reach the encoder, the transformer-based encoder, and the chart-style decoder.

\subsubsection{The speech-processing CNN}

Of the four prosodic features, pause and duration are already discrete at the token level. 
Pitch and intensity, however, are extracted from frames every 10 ms in the original speech signal. 
If a given token is shorter than a fixed number of frames, some frames of left and right context are included; frames from longer tokens are subsampled to reduce their frame length.
These two frame-based features features have a different dimensionality than the token-level input and they are untenably long for a sequence model or transformer. 
The CNN solves both these problems by producing a fixed-length representation for each feature at the token level. 
This representation can be concatenated with the other token-level features and input to the encoder.

For a speech input with $f$ frames, the raw features input to the CNN have dimensions $6 \times f$, where 6 is the number of total features for each frame (3 pitch features and 3 intensity features). 
Several filters of different sizes then perform one-dimensional convolution of the input. 
These different filters allow the CNN to integrate information on various time scales. 
We apply $N$ of each of these $m$ filters, for a total of $mN$ filters. 
We use the hyperparameters described by \newcite{tran2018}: $N$ = 32 filters of widths $w$ = [5, 10, 25, 50], for a total of $mN$ = 128 filters. 
The output of each filter is then max-pooled, which converts the features for a given token to a uniform dimension.

These CNN-processed features are then concatenated with the token-level prosodic features (pause and duration) and the text embedding for the token, and then input to the encoder. 
The CNN is trained along with the encoder-decoder model.

\subsubsection{The encoder}

The encoder is a standard transformer with eight attention heads, based on the work of \newcite{kitaev2018}. 
For each word of input $x_i$, the transformer encoder produces a representation of the forward context, $\overrightarrow{y_i}$, and the backward context $\overleftarrow{y_i}$. 
We represent a given span between indices $i$ and $j$ by subtracting the forward representations and backward representations and concatenating the results: 
\[
    v_{(i,j)} = [\overrightarrow{y_j}-\overrightarrow{y_i};\overleftarrow{y_j}-\overleftarrow{y_i}]
\]
The next section explains how we use this span representation $v_{(i,j)}$ to generate scores for constituents in a tree.

\subsubsection{The decoder}

The decoder is a chart-style span-based decoder. 
Its goal is to output the correct tree $T$ for an input $x_1,...,x_n$. 
Each tree's score $S(T)$ is simply the sum of the scores of its constituents, where each constituent is defined by a start index $i$, an end index $j$, and a label $l$. 

\[
    S_{tree}(T) = \sum_{i,j,label \in T}{S_{label}(i, j, l) + S_{span}(i, j)}
\]

As this formula for tree score shows, each constituent's score is made up of a label score and span score. 
Conceptually, the span score corresponds to the probability that a constituent exists that exactly covers span $(i,j)$ in the input; the label score reflects the probability that the span $(i,j)$ has a given constituent label (e.g., \textsc{s, np}). 
The decoder must have a way of determining the label score and span score for each constituent.

The label scores are generated by passing the span representation $v_{(i,j)}$ through a two-layer feed-forward network like the feed-forward networks \newcite{vaswani2017} use: 
\[
    FFN(x) = W_2(relu(W_1x + b_1)) + b_2 
\]
Following \newcite{kitaev2018}, we also include a layer normalization step ($LNorm$). 
This feed-forward network produces a vector for each span $S_{label}(i,j)$ whose size is the number of possible labels:

\[
    S_{label}(i,j) = M_2(relu(LNorm(M_1v_{(i,j)}) + c_1)) + c_2
\]

The $l$th element of this vector is the score for the label $l$:
\[
    S_{label}(i,j,l) = [S_{label}(i,j,)]_l
\]

We also need to calculate the span score, but calculating the score for all spans $(i,j)$ would be prohibitively inefficient. 
Instead, \newcite{kitaev2018}, following the approach of \newcite{stern2017} and \newcite{gaddy2018}, use a dynamic programming strategy based on the CKY algorithm. 
The score for a span $(i,j)$ is calculated in terms of the scores of its subspans, which allows span scores to be built up recursively from the stored scores of smaller spans. 
A given span $(i,j)$ can be split at any internal point into two subspans, $(i,k)$ and $(k,j)$. 
Each of these possible splits $(i,k,j)$ is assigned a score, calculated by summing the span scores of the subspans:
\[
	S_{split}(i,k,j) = S_{span}(i,k) + S_{span}(k,j)
\]
Then, to find the best score for this span $(i,j)$, we find the label and split that maximize the following sum:
\[
S_{best}(i, j) = \max_{l,k}[S_{label}(i, j, l) + S_{split}(i, k, j)] 
\]
All spans are recursively split into subspans, eventually arriving at single-word spans. Since there are no splits possible for a single-word span, the score for a single word span is simply that word’s best label score:
\[
	S_{best}(i,i+1) = \max_l [S_{label}(i,i+1,l)]
\]
This method requires that the grammar be in Chomsky-Normal form, which the model achieves by collapsing strings of unary rules and using dummy nodes to make $n$-ary rules into binary rules.

With this method of generating tree scores from span representations, we can then define the hinge loss for our predicted tree $\hat{T}$ compared to the gold tree $T*$, where $\Delta$ represents the Hamming loss on labeled spans:

\begin{multline} \nonumber
Loss(\hat{T},T*)  = \\
   \indent\indent \max[0,\max_T[\Delta(\hat{T},T*) + S_{tree}(\hat{T})] \\
   \indent - S_{tree}(T*)]
\end{multline}
    
We then use this loss function to train our encoder-decoder, including the CNN input module for speech.

\subsection{Model training details} \label{app:hparams}

We used the same training-development-test split of the corpus as \citet{tran2018} and \cite{tran2019}.
We used the hyperparameters specified in \cite{tran2019}'s code base, documented in Table \ref{tab:hparams}.
Each model was trained for 50 epochs on a single Nvidia GTX 1080 GPU, which took approximately 7 hours per model.
The text-only models have approximately 23M trainable parameters each, while the text+prosody models have approximately 20M trainable parameters.

\begin{table}[ht]
    \centering
    \begin{tabular}{lc}
    \hline
       \textbf{Hyperparameter}  & \textbf{Value} \\    \hline
    Epochs & 50 \\
    Text embedding dim. & 300 \\
    Max. seq. length & 270 \\
    Dropout & 0.3 \\
    Num. layers & 4 \\
    Num. heads & 8 \\
    Model dim. & 1536 \\
    Key/value dim. & 96\\
      \hline
    \end{tabular}
    \caption{Model hyperparameters. Note that the maximum sequence length for the SU-based model is 200 tokens.}
    \label{tab:hparams}
\end{table}

\subsection{Modifying the end-to-end model} \label{app:dummywt}

For the end-to-end model, we made one modification to the parser in order to try to boost its segmentation score.
The end-to-end model tends to produce trees with a single top node branching into many children, and since we consider each direct child of the top \textsc{turn} node to be an SU predicted by the end-to-end model, this leads to very low segmentation scores.
We hypothesized that we could improve the end-to-end model's SU segmentation score by discouraging this kind of highly-branching top node.
The parser is only capable of producing binary branching nodes, so in order to produce multiply branching nodes, the parser predicts a series of binary branching \textsc{dummy} nodes in its initial prediction.
In a post-processing step, each dummy node is removed and its children are attached the dummy node's parent to produce the final parse.
In order to discourage oversegmentation, we experimented with weighting \textsc{dummy} nodes with a greater loss penalty. 
We found that adding a weight of 0.5 for each \textsc{dummy} node led to improved parse and SU segmentation quality. 
Table \ref{tab:dummy} shows the results with various weights.
While higher weights continue to improve segmentation quality, particularly segmentation precision, as expected, they start to harm parse quality.
Note that the values in Table \ref{tab:dummy} are for just one random seed, while the values in Table \ref{tab:parseresults} are the mean of each score over five randomly seeded models; this is why the scores do not match exactly.

\begin{table}[h]
    \centering
    \begin{tabular}{llcccc}
    \hline
    & & \multicolumn{3}{c}{Segmentation} & Parse \\
    & Plt. & Prec. & Rec. & F1 & F1\\
 \hline
       \multirow{3}{*}{Txt} & 0 & 55.01 & 75.78 & 63.74 & 86.09 \\
        & 0.5 & 59.66 & 69.59 & 64.24 & \textbf{85.33} \\
        & 1 & 70.46 & 64.03 & 67.09 & 84.74 \\
\hline
      \multirow{3}{*}{Txt+pros} &  0 & 54.79 & 77.59 & 64.22 & 86.47 \\
        & 0.5 & 63.60 & 75.12 & 68.88 & \textbf{86.48} \\
        & 1 & 73.10 & 69.67 & 71.35 & 86.00 \\

 \hline
    \end{tabular}
    \caption{Development set SU segmentation precision, recall, and F1 scores, and parse F1 for the end-to-end model, given different penalties (plt.) for the dummy node. These results are reported for one random seed only, instead of five.}
    \label{tab:dummy}
\end{table}

\subsection{Reducing error propagation in the pipeline model} \label{app:pipe}

In order to reduce the effects of error propagation from the segmentation step to the parse step, we performed a post-processing step on the trees produced by the pipeline parser so that the model does not have to predict a node dominating every predicted SU (e.g., the \textsc{S} node shown in Figure \ref{fig:decorrel}\subref{fig:pipe}). 
The effect of this post-processing step on the overall scores is negligible, suggesting that even when the pipeline model doesn't have to predict a node dominating each predicted SU, it tends to do so.

When training the parser step of the pipeline, we use the sentence segmentations predicted by the first step.
When these segmentations result in an SU that isn't a fully-connected tree, we insert a node labeled \textsc{blank} into the tree dominating that SU.
This allows the parser to predict a \textsc{blank} node in cases where an SU shouldn't have a top node.
When the parser trained on this data predicts a \textsc{blank} node, we remove this node in post-processing, essentially allowing the parser to fail to predict an SU where the segmentation step predicts one.

\end{document}